\documentclass[10pt,twocolumn,letterpaper]{article}

\usepackage{cvpr}              % To produce the CAMERA-READY version
\usepackage{graphicx}
\usepackage{amsmath}
\usepackage{amssymb}
\usepackage{booktabs}
\usepackage{multirow}
\usepackage[table]{xcolor}
\usepackage{algorithm, float}
\usepackage{algpseudocode}

\usepackage{footmisc}
\usepackage{bm}
\usepackage{xspace}
\usepackage{mathtools}
\def\vdelta{{\bm{\delta}}}

\def\vg{{\bm{g}}}
\def\vh{{\bm{h}}}

\def\vx{{\bm{x}}}

\def\mA{{\bm{A}}}

\def\mD{{\bm{D}}}

\DeclareMathAlphabet{\mathsfit}{\encodingdefault}{\sfdefault}{m}{sl}
\SetMathAlphabet{\mathsfit}{bold}{\encodingdefault}{\sfdefault}{bx}{n}

\def\sR{{\mathbb{R}}}

\newcommand{\Ls}{\mathcal{L}}

\DeclarePairedDelimiter\ceil{\lceil}{\rceil}
\DeclarePairedDelimiter\floor{\lfloor}{\rfloor}

\makeatletter
\@namedef{ver@everyshi.sty}{}
\makeatother
\usepackage{pgfplots}
\pgfplotsset{compat=1.7}
\usepackage{tikz}
\usepackage[pagebackref,breaklinks,colorlinks]{hyperref}

\usepackage[capitalize]{cleveref}
\crefname{section}{Sec.}{Secs.}
\Crefname{section}{Section}{Sections}
\Crefname{table}{Table}{Tables}
\crefname{table}{Tab.}{Tabs.}

\def\cvprPaperID{1450} % *** Enter the CVPR Paper ID here
\def\confName{CVPR}
\def\confYear{2022}

\newcommand{\sampling}{\textsc{SBFS}}
\newcommand{\model}{\textsc{Locformer}}
\newcommand{\OOM}{OOM}

\newcommand*\rot{\rotatebox[origin=c]{90}}
\newcommand{\CC}[1]{\cellcolor{gray!15}}

\newtheorem{proposition}{Proposition}

\begin{document}

%%%%%%%%% TITLE
\title{\model{}: Enabling Transformers to Perform Temporal Moment Localization on Long Untrimmed Videos With a Feature Sampling Approach}

\author{Cristian Rodriguez-Opazo$^{*  \spadesuit}$ \qquad Edison Marrese-Taylor$^{* \clubsuit \dagger}$ \qquad Basura Fernando$^{\blacklozenge}$\\ \qquad  Hiroya Takamura$^{\clubsuit}$ \qquad  Qi Wu$^{\spadesuit}$\\
{\tt\small cristian.rodriguezopazo@adelaide.edu.au}\\
${}^{\spadesuit}$ University of Adelaide, Australian Institute for Machine Learning\\
${}^{\clubsuit}$ National Institute of Advanced Industrial Science and Technology (AIST) \\  ${}^{\dagger}$The University of Tokyo \qquad ${}^{\blacklozenge}$ A*STAR Singapore
}

\maketitle

%%%%%%%%% ABSTRACT
    %First, we propose a sampling technique that splits the input feature sequence into a fixed number of sections and selects a single feature per section using a stochastic approach, allowing us to obtain a feature sample-set that is representative of the video content for the task at hand, while also keeping the memory footprint constant.
    % (coined \sampling{}) 
\begin{abstract}
    We propose \model{}, a Transformer-based model for video grounding which operates at a constant memory footprint regardless of the video length, i.e. number of frames. \model{} is designed for tasks where it is necessary to process the entire long video and at its core lie two main contributions. First, our model incorporates a new sampling technique that splits the input feature sequence into a fixed number of sections and selects a single feature per section using a stochastic approach, which allows us to obtain a feature sample-set that is representative of the video content for the task at hand while keeping the memory footprint constant. Second, we propose a modular design that separates functionality, enabling us to learn an inductive bias via supervising the self-attention heads, while also effectively leveraging pre-trained text and video encoders. We test our proposals on relevant benchmark datasets for video grounding, showing that not only \model{} can achieve excellent results including state-of-the-art performance on YouCookII, but also that our sampling technique is more effective than competing counterparts and that it consistently improves the performance of prior work, by up to 3.13\% in the mean temporal IoU, ultimately leading to a new state-of-the-art performance on Charades-STA.
    % on the mean temporal IoU metric in three different dataset by a margin between 0.1\% and 3.13\%.
    % , but that it also be combined with our Transformer-based model to achieve the new state of the art.
    % \BF{I think better to mention numbers here.. what sort of improvement we get?}
     {\let\thefootnote\relax\footnote{{* Authors contributed equally}}}
\end{abstract}
%%%%%%%%% BODY TEXT

\section{Introduction}

%This task aims to identify the two temporal boundaries of a moment of interest based on an input untrimmed video, and a query in natural language \cite{richard2018neuralnetwork,lin_single_2017,escorcia_daps:_2016,Chao2018,GaoYSCN17,xu2019multilevel}. 

Vision-and-language understanding is an important problem which has largely attracted the attention from both the computer vision (CV) and natural language processing (NLP) communities. Among tasks in this area, the temporal localization of moments % also known as video grounding,
remains a fundamental and challenging one. This task aims to identify the start and end time of \emph{a moment of interest} in an input untrimmed video given the query in natural language \cite{richard2018neuralnetwork,lin_single_2017,escorcia_daps:_2016,Chao2018,GaoYSCN17,xu2019multilevel}. 
Recent approaches have aimed at directly predicting the starting and ending temporal locations, or regressing them from the input video \cite{yuan2018find,ghosh2019excl,rodriguez2019proposal}. Although these models are more efficient than previous propose-and-rank-based approaches, they still require a considerable amount of computation and/or memory since they need to process the whole input video at once. %\mr{to me it's not completely obvious (if I hadn't seen you struggling with CUDA errors) that videos are a big issue when it comes to memory. might be better to explicitly state this somehow, unless it's so obvious that it's stupid}

% \begin{figure}[t]
%     \centering
%     \includegraphics[width=0.5\textwidth]{figures/plot.pdf}
%     \caption{Overview of our method}
% \end{figure}
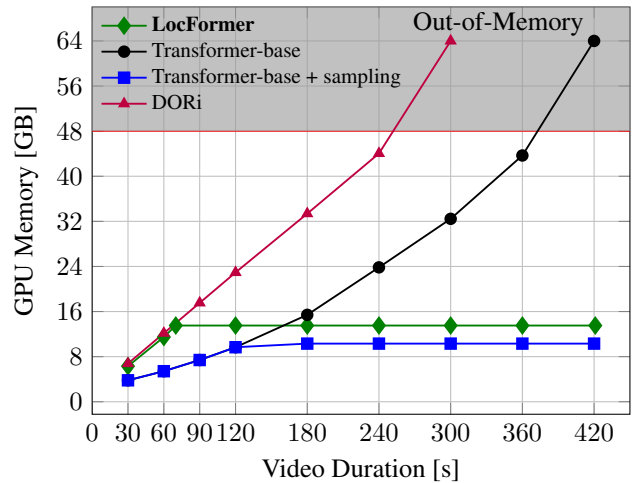
\begin{figure}[t!]
	\centering
\begin{tikzpicture}
\begin{axis}[
    grid=both,
    enlarge x limits=0,
    legend cell align={left},
    % legend style={font=\footnotesize},
    % label style={font=\footnotesize, text left},
    % legend columns=2, 
	xlabel={Video Duration [s]},
	ylabel={GPU Memory [GB]},
	width=0.5\textwidth,height=7cm,
    legend style={at={(0,1)},fill=none, draw=none,nodes={scale=0.8, transform shape},anchor=north west},
    xtick={0,30,60,90,120,180,240,300,360,420},
    ytick={0,8,16,24,32,40,48,56,64,72},]
\addplot[color=red] coordinates {
	(0, 48)
	(450, 48)
};
% \fill[blue!40!white, opacity=0.2] (5.57,-100) rectangle (194,443);
% \fill[blue!40!white, opacity=0.2] (5,-100) rectangle (19,443);
\fill[black!40!white, opacity=0.6] (0,700) rectangle (500,443);
% \draw [dashed] (31.4,-10) -- (31.4,700);
% \addplot[line width=0.25mm,color=purple, mark=triangle*, mark size=3pt] coordinates {
% 	(30, 6.311)
% 	(60, 11.48)
% 	(90, 19.37)
% 	(120, 30.02)
% 	(180, 64)
% };
\addplot[line width=0.25mm,color=black!50!green,mark=diamond*, mark size=3pt] coordinates { % LocFormer
	(30, 6.311)
	(60, 11.48)
	(70, 13.52)
	(120, 13.52)
	(180, 13.52)
	(240, 13.52)
	(300, 13.52)
	(360, 13.52)
	(421, 13.52)
};
\addplot[line width=0.25mm, color=black,mark=*, mark size=2pt] coordinates { % Transformer-Base
	(30, 3.805)
	(60, 5.422)
	(90, 7.404)
	(120, 9.674)
	(180, 15.418)
	(240, 23.840)
	(300, 32.448)
	(360, 43.682)
	(420, 64)
};

\addplot[line width=0.25mm,color=blue, mark=square*] coordinates { % Transformer-base + sampling
	(30, 3.805)
	(60, 5.422)
	(90, 7.404)
	(120, 9.674)
	(180, 10.314)
	(240, 10.314)
	(300, 10.314)
	(360, 10.314)
	(420, 10.314)
};
\node[] at (axis cs: 340,67) {Out-of-Memory};
\addplot[line width=0.25mm,color=purple, mark=triangle*, mark size=2pt] coordinates { %DOri
	(30, 6.781)
	(60, 12.158)
	(90, 17.54)
	(120, 22.93)
	(180, 33.368)
	(240, 44.044)
	(300, 64)
};
% \addplot[line width=0.25mm,color=black!50!green, mark=diamond, mark size=2pt] coordinates {
% 	(30, 6.781)
% 	(60, 12.158)
% 	(70, 13.276)
% 	(120, 13.276)
% 	(180, 13.276)
% 	(240, 13.276)
% 	(300, 13.276)
% 	(360, 13.276)
% 	(419, 13.276)
% };
\legend{, \textbf{LocFormer}, Transformer-base, Transformer-base + sampling, DORi}%, DORi + SBFS}
\end{axis}
\end{tikzpicture}
\caption{Empirical upper-bound memory consumption of \model{}, Transformer-base, with and without enabling our proposed sampling technique (\sampling{}), and DORi \cite{rodriguez-opazoDORiDiscoveringObject2021}. Memory usage (y-axis) is computed on an NVIDIA RTX-8000 GPU, using a batch size of 32 and assuming all sequences have maximum length for the given video duration. Each sample is created using synthetic video features that mimic I3D (x-axis). As seen, our sampling strategy leads to a constant memory footprint regardless of the video duration.}
\label{figure:memory_consumption}
\vspace{-0.6cm}
\end{figure}
% \begin{figure}[t]
%     \begin{tikzpicture}
%     	\begin{axis}[
%     		xlabel=Cost,
%     		ylabel=Error]
%     	\addplot[color=red,mark=x] coordinates {
%     		(2,-2.8559703)
%     		(3,-3.5301677)
%     		(4,-4.3050655)
%     		(5,-5.1413136)
%     		(6,-6.0322865)
%     		(7,-6.9675052)
%     		(8,-7.9377747)
%     	};
%     	\end{axis}
%     \end{tikzpicture}
%     \caption{Overview of our method}
% \end{figure}
% its\mr{their} 
Recent improvements in both NLP and CV tasks can be attributed to the Transformer \cite{Vaswani_NIPS_2017} model. Despite their success, one main drawback of these models is their computational cost, being a greatly limiting factor for many, with memory usage ballooning as model sizes increase to attain better performance. This issue has had a significant impact on the NLP community, recently leading to the proposal of several model variations to better deal with longer inputs such as documents \cite{beltagyLongformerLongDocumentTransformer2020, wang2020linformer, kitaev2020reformer}
%\mr{add reformer, linformer, sparse attention papers},
by using restricted, simplified or memory-efficient versions of the attention component. The advent of new, deeper Transformer-based models for image and video understanding suggests that these problems are likely to become more relevant in the CV community as well.
This is exemplified by many existing developments in Transformer models for video understanding only being capable of processing short inputs at a time due to memory constraints. 
For example, TimeSformer-large \cite{bertasiusSpaceTimeAttentionAll2021} can only process inputs that are 24 seconds long\footnote{TimeSformer-base/large reads 8 frames at 32/4 fps respectively.}, ViVIT \cite{arnabViViTVideoVision2021} and X-ViT \cite{bulatSpacetimeMixingAttention2021} can receive inputs that are up to 32 frames long, while other approaches like MERLOT \cite{zellersMERLOTMultimodalNeural2021} and ClipBERT \cite{leiLessMoreClipBERT2021a} specifically sample a single or a few frames from the whole input video. Recent Transformer-based models for video grounding have taken a different approach, instead choosing to aggregate the video features during pre-processing with pooling techniques \cite{zhangMultiStageAggregatedTransformer2021} in order to be able to handle longer inputs. 
% \mrminor
Furthermore, many of the aforementioned approaches rely on the split-and-aggregate approach when processing long videos, where inputs are divided into sections which are processed separately by the model. Though many of the downstream tasks considered by these models do not specifically require temporal reasoning over the input videos, their applicability to tasks requiring temporal reasoning is limited, as it keeps models from capturing interactions across segments. 
In light of this, we present \model{}, a Transformer-based model for the task of temporal moment localization which operates at a constant memory footprint regardless of the input length, as shown in Figure \ref{figure:memory_consumption}. The success of the \model{} relies on two key ideas. Firstly, \model{} incorporates Stochastic Bucket-wise Feature Sampling (\sampling{}), which splits the sequence of input video feature into a fixed number of buckets and selects a single feature per bucket per iteration using a stochastic approach during training. While bucketing enables us to keep the memory budget limited by effectively shortening the input sequence length to the number of buckets, the stochastic nature of our approach allows us to obtain a better coverage of the video with sufficient stochasticity, obtaining a feature sample-set that is representative of the video content for the task at hand. This allows us to attain better generalization than traditional sampling methods like video-level down-sampling, i.e. extracting frames at low frame-rate and feature-level pooling.
% \mr{examples of traditional sampling methods + do you compare with this? If not then it would be better mention the specific sampling method (e.g. fixed sampling?)}

Secondly, we propose a modular design that separates the functionality inside the model, enabling us to learn an inductive bias via supervising the behavior of some self-attention heads, but without interfering with the functionality of the heads of existing pre-trained models such as BERT \cite{Devlin_2018_BERT}, which we also incorporate into the model. Finally, we introduce a new loss to induce the correct temporal order of the predicted starting and ending locations of the target moment.

To demonstrate the effectiveness of our proposals, we conduct experiments on three challenging datasets, Charades-STA \cite{Gao_2017_ICCV}, ActivityNet Captions \cite{caba2015activitynet,Krishna_2017_ICCV} and YouCookII \cite{ZhXuCoCVPR18,ZhLoCoBMVC18}. We show that \model{} is able to obtain state-of-the-art performance in the latter, and competitive results elsewhere. Moreover, we also show how \sampling{} can be easily combined with prior work, improving their performance in all datasets, leading to a new state-of-the-art on Charades-STA. %\EMT{and YoucookII}. \mr{maybe you can make up a metric like perf-to-memory ratio or something like that and solidly claim that you massively outperform baselines? also, it would be nice to have a figure like Figure 1, but with performance on one axis---really sends the point home, something like \url{https://arxiv.org/pdf/2108.12409.pdf} fig. 1} 
We believe our results highlight the importance of sampling techniques as a valid mechanism to obtain better coverage of long input videos while keeping memory usage under budget. This ultimately provides a concrete direction for further research on tasks where it is necessary to cover long untrimmed videos, which include but are not limited to video grounding. %We will release our code and data to encourage future research in this area.

% \begin{enumerate}
%     %  generic \QW{This is risky. We only tested on video-language, right?}
%     \item We propose \sampling{}, a video sampling technique which is model and feature agnostic which can successfully improve performance of existing models while keeping their memory footprint constant.
%     \item We introduce \model{}, a full Transformer-based model for the task of temporal moment localization. This model (),  whch while leveraging existing pre-trained models
%         \item We propose a new loss function to induce the correct temporal order of starting and ending locations of the target moment

% \end{enumerate}

\section{Related Work}
% Our work is related to video grounding, which originates from the task of temporal action localization. The goal of this task is to solve the problem of recognizing and determining temporal boundaries of action instances in videos, with extensive previous work devoted to it \cite{Shou_2016_CVPR,guAVAVideoDataset2018,girdharVideoActionTransformer2019}.
% Our work is related to video grounding, which originates from the task of temporal action localization. The goal of this task is to solve the problem of recognizing and determining temporal boundaries of action instances in videos, with extensive previous work devoted to it \cite{Shou_2016_CVPR,guAVAVideoDataset2018,girdharVideoActionTransformer2019}.
% The goal of this task is to solve the problem of recognizing and determining temporal boundaries of action instances in videos, with extensive previous work devoted to it \cite{Shou_2016_CVPR,guAVAVideoDataset2018,girdharVideoActionTransformer2019}. 

\begin{figure*}
    \centering
   \includegraphics[width=0.9\linewidth]{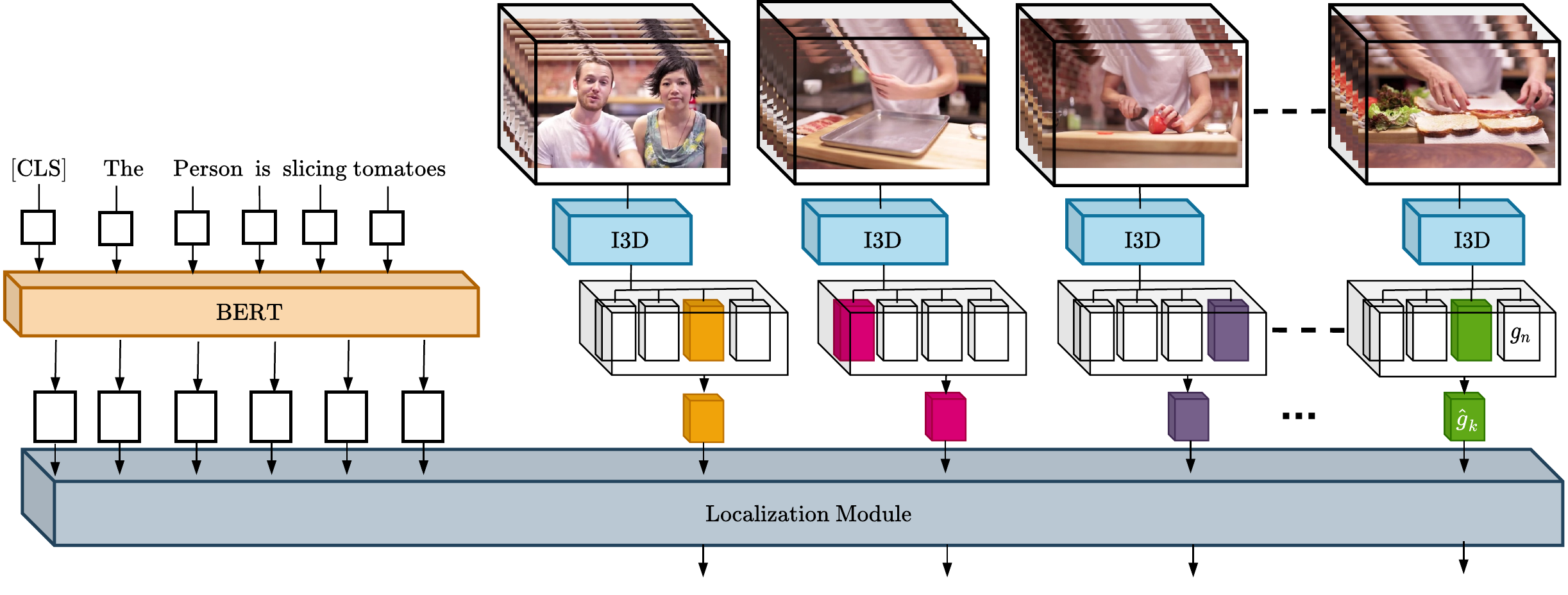}
    \caption{Our method uses a sampling technique that divides the video into a fixed number of buckets. Then, we uniformly sample a single I3D feature from each bucket, which is then fed into the localization module, a multi-modal Transformer model that receives video features and language features obtained from BERT.}
    \label{fig:overview}
    \vspace{-0.5cm}
\end{figure*}
\noindent \textbf{Temporal Action and Moment Localization.} The goal of \textit{Temporal Action localization} is to solve the problem of recognizing and determining temporal boundaries of action instances in videos, with extensive previous work devoted to it \cite{Shou_2016_CVPR,guAVAVideoDataset2018,girdharVideoActionTransformer2019}. Given the limitations of temporal action localization, which is restricted to a pre-defined list of labels, the task of language-driven \textit{temporal moment localization} was introduced as a generalization \cite{Gao_2017_ICCV,Hendricks_2017_ICCV}. In this task, the goal is to determine the start and end times of the video segment that best corresponds to a given natural language query. As this requires the model to extract useful information from the textual semantics in the query in order to identify the moment, this task is also usually regarded as video grounding. Early approaches that tackled temporal moment localization, including the work of Liu \etal. \cite{liu2018attentive} and Ge \etal. \cite{ge2019mac}, were mainly based on the generation of proposals or candidate clips which could later be ranked. Soon after, Chen \etal. \cite{chen-etal-2018-temporally}, Chen and Jiang \cite{sap2019}, and Xu \etal. \cite{xu2019multilevel} focused mainly on reducing the number of proposals by producing query-guided or query-dependent approaches. Recently, Zhang \etal. \cite{zhangMultiStageAggregatedTransformer2021} also adopted a Transformer-based model for this setting, being the most relevant to our work. % or on completely eliminating the proposal.

The extensive computation of enumerating candidates in the above mentioned proposal-based methods led to the development of methods that can directly output the temporal coordinates of the segment, namely, proposal-free approaches. In this context, Ghosh \etal. \cite{ghosh2019excl} first focused directly on predicting the start and end frames using regressions, and soon after Rodriguez-Opazo \etal. \cite{rodriguez2019proposal} improved results by modelling label uncertainty. While Mun \etal. \cite{munLocalGlobalVideoTextInteractions2020} and Zeng \etal. \cite{zengDenseRegressionNetwork2020} later proposed more sophisticated modality matching strategies, some more recent approaches have focused on better contextualizing I3D \cite{carreira_CVPR_2017_i3d} video features by proposing other model variations \cite{liProposalFreeVideoGrounding2021}. More recently, Liu \etal. \cite{liuAdaptiveProposalGeneration2021}, CPNet \cite{liProposalFreeVideoGrounding2021}, VSLNet \cite{zhangSpanbasedLocalizingNetwork2020a} have pushed performance further up. Finally, models like DORi, which also incorporates spatial features, \cite{rodriguez-opazoDORiDiscoveringObject2021} and CPN \cite{zhaoCascadedPredictionNetwork2021} have proposed ad-hoc graph-based approaches, with excellent results. Our contributions are orthogonal to these models, but also complementary because of their large memory consumption, as we will show in \S\ref{sec:sampling_combined}.
% but evaluations do not seem 100\% compatible with our approach as it is proposal-based

%Finally, Zhang et al. \cite{zhangWhereDoesIt2020} have recently proposed a novel task which requires not only to perform temporal language-driven moment localization, but also to spatially locate the objects mentioned in the query. Their approach is similar to ours in the sense that it also utilizes a spatio-temporal graph. However, the textual clues are incorporated after the graph construction rather than being an explicit part of it.

% \EMT{Here potentially talk about stochastic sampling aka monte carlo methods, mainly used for inference of intractable functions.}
\noindent \textbf{Sampling.} %Our feature-level sampling technique is also related to generic use of sampling in machine learning.
To the best of our knowledge, the earliest example of a sampling technique that is similar to ours is the work of Nakagawa \etal. \cite{nakagawaSpeaker-independent1988}, who proposed a stochastic version of dynamic time warping for speech recognition of the Japanese language in the late 80s. This idea was further extended by Chendra \etal. \cite{chendraRandomized2016} in the context of motion recognition, where a randomized version of dynamic time warping for this task was introduced. We also find several models that utilize sampling techniques for action recognition in videos, in this case to specifically select salient clips, such as SCSampler \cite{Korbar_2019_ICCV} and MGSampler \cite{zhiMgsampler2021}, which were later adopted by models like MVFNet \cite{wu2020MVFNet}. Also, Adaframe \cite{Wu_2019_CVPR} recently proposed a framework that adaptively selects relevant frames on a per-input basis for fast video recognition.

% \paragraph{Sampling in Computer Vision}
% \begin{itemize}
    % \item \cite{wu2020MVFNet} Mvfnet: Multi-view fusion network for efficient video recognition
    % \item Multi-agent reinforcement learning based frame sampling for effective untrimmed video recognition.
    % \item \cite{Korbar_2019_ICCV} SCSampler: Sampling Salient Clips From Video for Efficient Action Recognition
    % \item \cite{zhiMgsampler2021} MGSampler: An Explainable Sampling Strategy for Video Action Recognition
% \item \cite{Wu_2019_CVPR}: Adaframe, a framework that adaptively selects relevant frames on a per-input basis for fast video recognition.
% \end{itemize}
\noindent \textbf{Transformers.}
Finally, our work is also related to Transformer models proposed in the context of video-and-language understanding, and natural language processing. While most existing work is concerned with the pre-training of models on large datasets using general-purpose tasks such as masked-language modelling \cite{chenUNITERUNiversalImageTExt2020, luViLBERTPretrainingTaskAgnostic2019}, our approach in this paper is more specific and directly trains the model for the task at hand. % add the transformer-based proposal-based method

\section{Proposed Approach}
% \QW{Structure is not good. Transformer $->$ sampling $->$ loss}
% \EMT{Hi Qi, thank you for pointing this this out! I have thought a lot about the structure of this section, and I agree that the current version is not very good. Chris and I are still running experiments to find some additional gains to help re-design this section, please wait. Now, regardless of what happens there one issue with reverting the order here: we have to be careful as some Transformer-based experiments simply cannot be performed without also using the sampling, because the model does not fit even our larger GPUs.}

\subsection{Overview}

We assume that a given video $V \in \mathcal{V}$ can be characterized as a sequence of frames such that $V = \{v_t\}$ with $t = 1, \ldots, l$. Each video in $\mathcal{V}$ is annotated with a natural language passage $S \in \mathcal{S}$ where $S$ is a sequence of tokens $S = \{s_j\}$ with $j = 1, \ldots, m$, which describes what is happening in a certain period of time. Formally, this interval is defined by $t^s$ and $t^e$, the starting and ending points of the annotations in time, respectively. Although in the data a given video may be annotated with more than one single moment, and one natural language description may be associated to multiple moments, in this work we assume each derived case as an independent, separate training example.

Our model is trained to predict the most likely temporal localization of the contents of a given input query $S$ in terms of its start and end positions $t^{s\star}$ and $t^{e\star}$ in the video. We apply the mapping $\tau = (t \cdot n \cdot \text{fps}) / l$ to transform frame/feature index to time, converting $t^s$ and $t^e$ into $\tau^s$ and $\tau^e$, which correspond to specific integer feature positions such that $\tau^s, \tau^e \in [1, \ldots, n]$.

% \mrminor 
\model{} follows the Transformer architecture \cite{vaswani2017attention}, which has been recently extended to multi-modal scenarios as in UNITER \cite{chenUNITERUNiversalImageTExt2020} in the context of vision-and-language, and Recurrent VLN-BERT \cite{hongVLNBERTRecurrent2021} for vision-and-language navigation. Our model operates on a sequences of tokens $\{s_j\}$ and a video $\{v_t\}$ characterized as a sequence of frames, as specified earlier. The overall architecture is composed by three main modules: (1) The Video Encoding Module, which is in charge of mapping video frames to vectors, and obtaining a sample that is representative of the video contents, (2) the Text Encoding Module, a Transformer model with dimension $d_m$ in charge of extracting useful representations from the natural language query, and (3) the Localization Module, a multi-modal Transformer, also with hidden dimension $d_m$, which receives both textual and video features from the previous modules and is in charge of estimating $\tau^s$ and $\tau^e$. In the following subsections, we give details about each component and how they interact.

%  \EMT{alternatively: Video sampling module}
\subsection{Video Encoding Module with Feature Sampling}

% We are interested in developing a technique to limit the memory budget of our transformer model a given model when dealing with long video inputs.

% For an original such input sequence of $n$ features $F = [\vf_1 \ldots, \vf_n]$, our sampling technique with a bucket size of $b$ works

Our video encoding module is in charge of mapping the $l$ input video frames into a sequence of video features $G =\{\vg_i \in \mathbb{R}^{d_v}\}$, $i=1, \ldots, n$, and selecting a subset of these features that is representative of the contents of the video, which will later be fed into the localization module.

% We assume input video features $G$ are extracted by a video encoding function $F_V(V)$, and propose to limit the overall memory budget of the model by shortening the sequence of video features fed into the localization module into a fixed length. We do this in practice by proposing a technique that we call Stochastic Bucket-wise Feature Sampling (\sampling{}), which returns a sequence of length $b$ derived from $G$, as follows.
% \begin{equation}
%     \sampling_b(F) =
%         \begin{cases}
%             \begin{matrix}
%                 \{\vg_i\}_{i=1}^n & \text{if} \; n\leq b \\
%                 \{\vg_{f(k)}\}_{k=1}^{m=n/b} & \text{if} \; n > b
%             \end{matrix}
%     \end{cases}
%     \label{eq:sampling}
% \end{equation}
% In Equation \ref{eq:sampling}, the index $f(k)$ is sampled according to a uniform distribution over the indices of the features in the bucket, as Equation \ref{eq:uniform} shows, below.
% \begin{equation}
%     f(k) \sim \mathcal{U}_{(n/b) \cdot k}^{(n/b) \cdot (k+1)}
%     \label{eq:uniform}
% \end{equation}
We assume input video features $G$ are extracted by a video encoding function $F_V(V)$, and propose to limit the overall memory budget of the model by shortening the sequence of video features fed into the localization module. We do this in practice by proposing a technique that we call Stochastic Bucket-wise Feature Sampling (\sampling{}), which returns a sequence of length at most $b$ derived from $G$, as follows.
\begin{equation}
    \sampling(G;b) =
        \begin{cases}
            \begin{matrix}
                \{\vg_i\}_{i=1}^n & \text{if} \; n\leq b \\
                \{\vg_{f(k)}\}_{k=1}^{m(n,b)} & \text{if} \; n > b
            \end{matrix}
    \end{cases}
    \label{eq:sampling}
\end{equation}
In Equation \ref{eq:sampling}, $m(n,b)$ characterizes the number of buckets to be allocated to host video features, and is defined as $m(n,b) = \floor{\frac{n}{\ceil{n/b}}} \leq b$, where $\floor{}$, $\ceil{}$ are the floor and ceiling operators, respectively. The index $f(k)$ is sampled according to a uniform distribution over the indices of the features in the bucket, as Equation \ref{eq:uniform} shows, below.
\begin{equation}
    f(k) \sim \mathcal{U}_{\ceil{n/b} \cdot (k-1)}^{\ceil{n/b} \cdot k}
    \label{eq:uniform}
\end{equation}
% In simple words \mr{More intuitively},
More intuitively, we create a fixed number of buckets and allocate features to each by equally distributing them into the buckets. During training, we randomly sample a single feature for each bucket, following a uniform distribution, effectively reducing the input sequence length to at most $b$, the number of buckets. When doing this, we also accordingly convert the original set of labels $\tau^s, \tau^e \in [1, \ldots, n]$ into $\bar{\tau}^s, \bar{\tau}^e \in [1, \ldots, b]$. For simplicity, without loss of generality, for the rest of the paper we will assume the sampling module always returns total of $b$ video features.
% \BF{Compared to random sampling of frames, this sampling strategy allows us to obtain a better coverage of the video with sufficient randomness that helps better generalization. Our sampling method is also related to Monte Carlo Simulation methods. Our objective is to get a good sample-set for a given video that is representative of the video content. Our strategy, is more likely to generate a sample-set that is more representative than random sampling. I just wonder we can formally prove this..}
% \EMT{Hi Basura, thank you for pointing these things out, they have helped me better contextualize the approach, and more clearly understand its nature and potential.}

% The nature of \sampling{}, where a single feature is sampled from each bucket with equal probability each time a given example is fed to the model, grants it a data augmentation quality, as the model is exposed to different example variations on each epoch. Concretely,
% \BF{Please check if this is okay.}
The nature of \sampling{} is that a single feature is sampled from each bucket with equal probability each time. 
If we assume features are i.i.d., this implies that the probability of getting a sampled video feature sequence is $P(\tilde{G}) = (b/n)^b$.
As this is a very small probability, the number of potential distinct sampled video feature sets ($\tilde{G}$) is exceptionally large.
% . \mr{hahaha ``trump intensifies''}
% $\sampling_b(F)$
% \mr{i can infer why but at a more intuitive level a short description would be nice}
Fortunately, video features (frames) within a bucket are highly correlated as the original from neighboring video frames which are generally very similar, and one may make a weak assumption that bucket population in bucket $k$ may not contain sufficiently more information than any sampled feature $\vg_{f(k)}$ from the bucket population. 
In other words, if $\vg_{f(k)}$ is sufficient statistic of the bucket $k$, then any sampled video feature sequence $\tilde{G}$ using \sampling{} contains sufficient statistics of $G$, and $\sampling(G)$ is the sufficient statistic of video feature population $G$. Therefore, we can make the following proposition. 
% \mr{I don't know, this feels quite a strong proposition to make. This finding may only be significant for temporal moment localization.... Maybe make the proposition more specific/less assertive?}
\begin{proposition}
Any sampled video feature sequence $\tilde{G}$ from \sampling{} method is a sufficient statistic of video feature population $G$.
\end{proposition}
% \mr{this should probably be reworded}
The above proposition is very important as it allows us to train any complex model, such as a Transformer, on very long videos using \sampling with adequate guarantees. Next, we also present an interesting insight on how to pool features within a bucket.
%For a sequence length of 2,000 and a total of 100 buckets, we can see that this evaluates to $P(V) = 7.8 \times 10^{131}$, which essentially means that if we want the exact sequence of video features to be sampled again, we have to rub $~10^{130}$ epochs. \EMT{Potentially mention the theoretical/monte carlo analysis here.}
%Note that our feature extractor $F_V(V)$ generates a sequence of feature vectors $\vg_1, \cdots, \vg_n$ and if we denote the $j-th$ dimension of vector $\vg_i$ by $g_i^j$.
%Now if we assume that the features generated by  extractor $F_V(V)$ is bounded, i.e. $g_i^j < \beta$ for any video for any $i$ and $j$, and uniformly distributed, 
To do this, let us denote the $j$-th dimension of the feature vector $\vg_i$ by $g_i^j$.
\begin{proposition}
If the $j$-th dimension of vectors within the bucket $k$ has a uniform population for all $j$, i.e. $g_{f(k)}^j$ is uniformly distributed
on $[1, \mu_j]$ where $\mu_j$ is unknown, by the Fisher–Neyman factorization theorem~\cite{fisher1922mathematical}, the sufficient statistics of the population within the bucket $k$ is given by the max-pooling operator over the bucket features \footnote{Please check the supplementary material for details.}.
% \mr{maybe make the supplementary part a footnote? it's not part of the proposition. not a big deal though} %\BF{We can give the proof in the supplementary}.
\end{proposition}

Although our \sampling{} procedure can also be applied during inference, founded by Proposition 2, it is better to decouple the model from this stochastic component and instead utilize max pooling operator over the features of the bucket at inference time. 
This gives models increased stability when predicting, without sacrificing performance, as we will show in \S\ref{sec:ablations}.

% \BF{Have we considered sampling K frames from a bucket and then applying max/average pooling}
% \EMT{I discussed a similar idea with Chis a while a go, but I haven't tried it yet.}

\subsection{Text encoding module}

In the text encoding module, sentences are split using the BERT tokenizer, which also prepends the special \texttt{CLS} token, and adds the \texttt{SEP} marker at the end. Each token is mapped to learned embeddings of dimension $d_m$ and summed with learned positional encodings of the same size. These vectors are passed through $\bar{L}$ encoder transformer blocks \cite{vaswani2017attention} with $\bar{M}$ attention heads, to produce final text representations $[\bar{\vh}_0, \dots, \bar{\vh}_{m}]$. 

% \EMT{Mention that this module, is, for this iteration, not trainable.}
% \BF{I think we have to tell people, this is just a feature extractor right? We don't train BERT....}

\subsection{Localization module}

As mentioned earlier, the localization module is a Transformer model that receives both textual and video features, previously obtained by the respective modules. For the former, we directly input $\bar{\vh}_0, \ldots, \bar{\vh}_{m}$, while for the latter we first project $\tilde{G} = [\tilde{\vg}_1, \ldots, \tilde{\vg}_b]$ into the hidden dimension using a trainable linear layer and further combine this with a set of learned positional encodings. These two encoded vector sequences are concatenated lengthwise and passed through $L$ encoder blocks \cite{vaswani2017attention} with $M$ attention heads, to produce $[\vh_0, \dots, \vh_{m+b}]$.

From these vectors, we select $[\vh_{m+1}, \dots, \vh_{m+b}]$ and utilize the same localization function proposed by \cite{rodriguez2019proposal} as the main training signal, namely, feature-level soft classification task on the time dimension. Concretely, two different MLP layers produce scores of each position being the start/end of the location, which are passed through a softmax activation to obtain $\hat{\boldsymbol {\tau}}^s, \hat{\boldsymbol{\tau}}^e \in \mathbb{R}^{b}$, which are compared to soft-labels using the Kullback-Leibler divergence ($\mathcal{L}_{KL}$).

In order to guide the model to utilize the information in the relevant section of the video, we encourage the attention heads of this module to put more weight into the target video portions during training, adapt the approach proposed by \cite{rodriguez2019proposal} as shown below.
% \begin{equation}
%     \mathcal{L}_{att} = - \sum_{i=1}^b \sum_{l=1}^L \sum_{m=1}^M (1- \delta_{\tau^s \leq i \leq \tau^e})^\intercal \log(1-\mA_{:,i}^{(l,m)}) %\frac{1}{B} \sum_{i=1}^B
%     \label{eq:att_focus}
% \end{equation}
% In Equation \ref{eq:att_focus}, $\delta$ is the Kronecker delta, returning 1 when $i$ is inside the range of $\tau$, is a row vector, and $\mA_{:,i}^{(l,m)}$ is the $i$-th column in the attention matrix of the $l$-th layer and $m$-th attention head of the localization module.
% About equation 3. I believe the best way to explain this is by creating a vector x with the kronecker delta which dimension is equal to the length of the input sequence that feed the localization module and the indices  of the moment and language are 1.
\begin{gather}
    \mathcal{L}_{att} = - \sum_{l=1}^L \sum_{m=1}^M  (\bm{1} -\vx \otimes \vx) * \text{log}(\bm{1} - \mA^{l,m})
    \label{eq:att_focus}
\end{gather}
% x = (1 - \delta_{\tau^s \leq i \leq \tau^e}) \in \mathbb{R}^{m+b} where i \in [0,...,m+b] then we compare the attention matrix with the outter product of x so x \otimes x - A^l, where A^l is the attention matrix on layer l. then
% \sum_l \sum_i \sum_j x \otimes x - A_l
In Equation \ref{eq:att_focus}, $\mA^{(l,m)}$ is the the attention matrix of the $l$-th layer and $m$-th attention head of the localization module and $\vx \in \mathbb{R}^{m+b}$ is a vector that denotes which areas of the output sequence will be subject to our guiding signal, and is defined as $\vx = [\bm{1}_m;\vdelta_{\tau^s \leq i \leq \tau^e}]$, where $;$ denotes concatenation, $\delta$ is the Kronecker delta returning 1 when $i$ is inside the range of $\tau$, and $\bm{1}_{k}$ denotes a vector of ones of size $k$.
% of the length of the query.

% when $i$ is inside the range of $\tau$, is a row vector, and 
% While this is reportedly \cite{rodriguez2019proposal} not an issue, we have empirically observed that it indeed leads to flipped predictions in long videos. %\BF{why it is not an issue? i would say this is a limitation of\cite{rodriguez2019proposal}.} \CRO{indeed this is a limitation of \cite{rodriguez2019proposal} and I think that is what we try to convey}
% we propose a temporal consistency loss to encourage the predicted starting location to occur before the predicted ending location. 
We also note that the localization loss proposed in \cite{rodriguez2019proposal} is not sensitive to the order of the predictions of the starting and ending locations, as there is no conditioning on the time in the model portions that generate them. To induce the model to respect this order, we take a probabilistic approach and push the expected of the start of the segment (S) to be before the expected value of the ending (E) location, which is equivalent to requiring $\mathbb{E}(E) - \mathbb{E}(S) > 0$. Replacing the values of the expectations, we obtain the following.
\begin{equation}
    \mathbb{E}(E) - \mathbb{E}(S) = \sum_{i=1}^b \hat{\tau}^e_i i - \sum_{i=1}^b \hat{\tau}^s_i i = \sum_{i=1}^b i (\hat{\tau}^e_i - \hat{\tau}^s_i)
    \label{equation:expected_se}
\end{equation}
In Equation \ref{equation:expected_se} above, $\hat{\tau}^s_i$ and $\hat{\tau}^e_i$ are integers that denote the predicted probability value of the starting and ending localizations at position $i$. Based on this derivation, we formally implement our loss by minimizing the negative difference of the expected values as shown in Equation \ref{equation:se_loss}, below.
\begin{equation}
    \mathcal{L}_{se} = \min(0, \sum_{i=1}^b i (\hat{\tau}^s_i - \hat{\tau}^e_i))
    \label{equation:se_loss}
\end{equation}
Finally, our model is trained with the direct summation of the three losses introduced earlier such that $\Ls =  \Ls_{KL} + \Ls_{att} + \Ls_{se}$.

% \BF{SO what is the total loss function?}

%VLN BERT \cite{hongVLNBERTRecurrent2021}
%         Model annotation uncertainty, we take $\tau^s$ and $\tau^e$ and create two target categorical distribution vectors $\boldsymbol{\tau}^s \sim \mathcal{N}(\tau^s$, $1) \in \mathbb{R}^{n}$ and $ \boldsymbol{\tau}^e \sim \mathcal{N}(\tau^e$, $1) \in \mathbb{R}^{n}$ using a quantized Gaussian distribution.
%             \begin{equation}
%                 L_{KL} = \displaystyle D_{\text{KL}}(\hat{\boldsymbol{\tau}}^s \parallel \boldsymbol {\tau}^s) + \displaystyle D_{\text{KL}}(\hat{\boldsymbol {\tau}}^e \parallel \boldsymbol {\tau}^e)
%                 \label{eq:kl_div}
%             \end{equation}

\section{Experiments}

\subsection{Datasets}
To evaluate our proposed approach, we work with three widely-utilized and challenging datasets.
%, namely Charades-STA \cite{Gao_2017_ICCV}, ActivityNet Caption \cite{caba2015activitynet,Krishna_2017_ICCV}, YouCookII  \cite{ZhXuCoCVPR18,ZhLoCoBMVC18} which was recently (utilized by \EMT{DORi}), and TACoS \cite{tacos}.
% In Gao et al.\cite{Gao_2017_ICCV}, the sentences describing the video were semi-automatically decomposed into smaller chunks and aligned with the activity classes, which were later verified by human annotators. As a result of this process, the original class-based activity annotations are effectively associated to their natural language descriptions, totalling 13,898 pairs.
\textbf{Charades-STA}: Built upon the Charades dataset \cite{sigurdsson2016hollywood}, which provides time-based annotations using a pre-defined set of activity classes, and general video descriptions. We use the predefined train and test sets, containing 12,408 and 3,720 moment-query pairs respectively. Videos are 31 seconds long on average, with 2.4 moments on average, each being 8.2 seconds long on average.
% In our ablation studies we randomly split the train set in 80\% for training and 20\% for evaluation of the experiments.
\textbf{ActivityNet Captions}: Introduced by Krishna \etal. \cite{Krishna_2017_ICCV}, this dataset originally constructed for dense video captioning, consists of 20k YouTube videos with an average length of 120 seconds. The videos contain 3.65 temporally localized time intervals and sentence descriptions on average, where the average length of the descriptions is 13.48 words. Following the previous methods, we report the performance on the combined validation sets.
\textbf{YouCookII}: Consists of 2,000 long untrimmed videos from 89 cooking recipes obtained from YouTube by Zhou \etal. \cite{ZhXuCoCVPR18}. Each step for cooking these dishes was annotated with temporal boundaries and aligned with the corresponding section of the recipe. The average video length is 5.26 minutes. In terms of relevant moment segments, each video has 7.73 moments on average, with each segment being 19.63 seconds long on average.
% Recipes are written following the usual style of the domain  \cite{linStyleVariationCooking,gerhardt2013culinary}, which includes very specific instruction-like statements with a wide degree of detail.
% The videos were taped by individual persons at their houses while following the recipes using movable cameras.
% Videos have a minimum of 3 and a maximum of 16 moments.

% \textbf{MPII TACoS}: built on top of the MPII Compositive dataset \cite{rohrbach2012script}, it consists of videos of cooking activities with detailed temporally-aligned text descriptions. There are 18,818 pairs of sentence and video clips in total  with the average video length being 5 minutes. A significant feature of this dataset is that due to the atomic nature of many of the descriptions ---e.g. ``takes out the knife`` and ``chops the onion''--- the associated video moments only span over a few seconds, with 8.4\% of them being less than 1.6 seconds long. This makes this dataset specially challenging for our task, as the relative brevity of the moments allows for a smaller margin of error. When it comes to splits, we use the same as in \cite{Gao_2017_ICCV}, consisting of 50\% for training, 25\% for validation and 25\% for testing.

\subsection{Implementation Details}
\label{section:implementation_details}

% TACoS
For our experiments, we consider an off-line video encoding function $F_V(V)$, following previous work~\cite{ghosh2019excl,rodriguez2019proposal,rodriguez-opazoDORiDiscoveringObject2021,wangTemporallyGroundingLanguage2019,yuanSemanticConditionedDynamic2019,yuan2019semantic,zhang2018man}. Concretely, we first pre-process the videos by extracting features of size $1024$ using I3D with average pooling, taking as input the raw frames of dimension $256 \times 256$, at 25fps. We use the pre-trained model trained on Kinetics for ActivityNet and YouCookII released by \cite{carreira2017quo}. For Charades-STA, we use the pre-trained model trained on Charades. For the natural language input, we use the BERT-base-uncased tokenizer and keep the parameters of the Text Encoder fixed. Our experiments are performed on two GPUs, a 16-GB NVIDIA V100 and a 48-GB Quadro RTX 8000. Models are trained in an end-to-end fashion using ADAM \cite{kingma_adam}.

% \EMT{Here consider a short explanation about why we don't consider other features and/or directly put some pre-trained model on top of ours}.
Evaluation is based on two widely used metrics proposed by \cite{Gao_2017_ICCV}, namely the Recall at various thresholds of the temporal Intersection over Union (tIoU or $R@\alpha$) measuring the percentage of predictions that have tIoU with ground truth larger than certain $\alpha$, and the mean averaged tIoU (mIoU). We use three $\alpha$ threshold values $0.3$, $0.5$ and $0.7$.

\subsection{Ablation Studies}
\label{sec:ablations}

We begin by performing an extensive empirical study of our proposed stochastic sampling technique, comparing it to several alternatives. We specifically consider the following sampling approaches.

\textbf{Random:} As a naive baseline, we randomly sample features from the video, maintaining the order.
% of the video sequence.

\textbf{Fixed-rate video down-sampling (FRVS)}: We experiment with I3D features extracted at a lower frame-rate of 5fps, a technique that has been utilized by previous work such as \cite{ghosh2019excl}, which can be regarded as a form of low-level down-sampling.

\textbf{Fixed-rate feature down-sampling (FRFS):} We experiment with two fixed-rate down-sampling techniques at the feature level, bucket-level mean-pooling and max-poling. 

\textbf{Dynamic Time Warping (DTW):} We perform dynamic time warping between the non-structured video features and the fixed size temporal sequence created using our stochastic sampling technique and max-pooling applied inside each bucket. In this way, we assign features to each bucket that will later be randomly sampled.

\textbf{Dynamic-rate feature down-sampling (DRFS):} We utilize the similarity across features to dynamically create each bucket. While many variations are possible here, we decided to utilize a cosine distance-based heuristic to create the buckets. Please check the supplementary material for details. %\EMT{Explain this in the supplementary material}.

\textbf{\sampling{} Variations:} Taking our proposed technique as a base, we experiment with different alternatives for inference. Concretely, we always apply our stochastic sampling during training, and either use it for inference as well (\sampling{}-all), or replace it with bucket-wise mean pooling (\sampling{}-mean) or max pooling (\sampling{}).

For the experiments, we combine each of these sampling approaches with the rest of the \model{} architecture, and always use a bucket size of 200. Regarding the data, we use the YouCookII datasets, as contains videos that can be as long as 18 minutes, and contains queries that use rich language, which should help illustrate the importance of the sampling more clearly.

% \begin{table}[ht]
%     \centering
%     \begin{tabular}{lllllllll}
%         \toprule
%         \multirow{2}{*}{Sampling} & \multicolumn{4}{c}{Anet Cap} & \multicolumn{4}{c}{YouCookII} \\
%         & 0.3 & 0.5 & 0.7 & mIoU & 0.3 & 0.6 & 0.7 & mIoU \\
%         \midrule
%         Random &  &  &  &  &  &  &  &  \\
%         Mean FRS &  &  &  &  &  &  &  &  \\
%         Max FRS &  &  &  &  &  &  &  &  \\
%         DTW &  &  &  &  &  &  &  &  \\
%         DRW &  &  &  &  &  &  &  &  \\
%         SS &  &  &  &  &  &  &  &  \\
%         SS + mean &  &  &  &  &  &  &  &  \\
%         SS + max &  &  &  &  &  &  &  &  \\
%         \bottomrule
%     \end{tabular}
%     \caption{Results of our sampling ablation?.}
%     \label{table:sampling_alternatives}
% \end{table}
% \usepackage{multirow}
\begin{table}[t]
    \centering
    \rowcolors{5}{gray!15}{white}
    \begin{tabular}{lrrrr}
        \toprule
        \multirow{2}{*}{\textbf{Sampling}} & \multicolumn{4}{c}{\textbf{Performance}} \\
        \cmidrule{2-5}
        & R@0.3   & R@0.6   & R@0.7   & mIoU  \\
        \midrule
        Random & 9.42 & 3.35 & 0.74 & 9.24 \\
        FRVS-5fps & 37.54 & 22.68 & 10.62 & 23.90 \\
        FRFS-mean & 45.99 & 31.01 & 15.46 & 29.98 \\
        FRFS-max & 45.53 & 30.61 & 15.55 & 30.11 \\
        DTW & 27.58 & 13.52 & 4.47 & 18.22 \\
        DRFS & 33.48 & 17.84 & 6.56 & 21.57 \\
        \sampling-all & 46.28 & 30.04 & 15.32 & 30.34 \\
        \sampling{}-mean & 46.68 & 30.61 & 15.23 & 30.53 \\
        \sampling & \textbf{46.76} & \textbf{31.33} & \textbf{15.81} & \textbf{30.92} \\
        \bottomrule
    \end{tabular}
    \caption{Performance of our model on the YoucookII dataset when we replace \sampling{} with alternative sampling techniques.}
    \label{table:sampling_alternatives}
\end{table}

        % Random & 9.31 & 2.69 & 0.43 & 8.14 \\ 
        % Mean FRS & 10.68 & 4.78 & 1.23 & 7.42 \\ 
        % Max FRS & 37.83 & 24.54 & 12.74 & 25.42 \\ 
        % DTW & 19.73 & 9.48 & 3.24 & 13.61 \\ 
        % DRS & 27.72 & 14.63 & 5.24 & 18.04 \\
        % SS & 39.29 & 25.40 & 12.80 & 26.20 \\ 
        % SS + mean & 10.22 & 4.73 & 1.32 & 6.58 \\ 
        % SS + max (\sampling{}) & \textbf{40.55} & \textbf{25.74} & \textbf{13.32} & \textbf{26.98}\\

As the results in Table \ref{table:sampling_alternatives} show, the effectiveness of our sampling technique is clear, specially when compared with more naive alternatives like random sampling, or simple mean pooling. We see that low-level down-sampling techniques that extract fewer frames from the original video, are not effective either. In contrast, the naive version of the max-pooling-based sampling stands out, performing similarly but still below \sampling{}. These results helps illustrate the importance of the stochastic approach we have taken, which enables us to limit the input to the model while still exposing it to all of the training data in the long run, significantly improving its generalization capabilities. Finally, we also note that all the tested sampling alternatives except DTW and DRFS do not utilize information about the features when generating the buckets. It is interesting to see that many of these arguably simpler sampling techniques, including \sampling{}, outperform data-informed approaches. 
% \EMT{What conclusion can we draw from this?}

\begin{table}[t!]  
    \centering
    \rowcolors{5}{gray!15}{white}
    \begin{tabular}{ccccc}
        \toprule
        \multirow{2}{*}{\textbf{Bucket Size}} & \multicolumn{4}{c}{\textbf{Performance}} \\
        \cmidrule{2-5} 
        & R@0.3 & R@0.6 & R@0.7 & mIoU \\
        \midrule
        100 & 35.74 & 27.41 & 11.88 & 29.03 \\
        200 & \textbf{46.76} & \textbf{31.33} & 15.81 & \textbf{30.92} \\
        300 & 46.36 & 31.07 & \textbf{15.95} & 30.63 \\
        400 & 45.25 & 30.30 & 15.58 & 30.23 \\
        500 & 44.33 & 28.69 & 14.18 & 29.14 \\
        \bottomrule
    \end{tabular}
    \caption{Impact on performance of \model{} on the YoucookII dataset as parameter $b$, the bucket size, changes.}
    \label{table:bucket_size_study}
\end{table}

\begin{table}[t]
    \centering
    \small
    % \rowcolors{5}{gray!15}{white}
    % \footnotesize
% \usepackage{booktabs}
% \usepackage{multirow}
%   \rowcolors{1}{}{gray!10}[respect-blocks]
\begin{tabular}{c@{\hspace{0.1cm}}c@{\hspace{0.1cm}}c@{\hspace{0.1cm}}c@{\hspace{0.1cm}}l@{\hspace{0.1cm}}l@{\hspace{0.1cm}}l@{\hspace{0.1cm}}l}
\toprule
\multirow{2}{*}{\textbf{Method}} & \multirow{2}{*}{\sampling{}} & \multirow{2}{*}{\textbf{$\mathcal{L}_{att}$}}  & \multirow{2}{*}{$\mathcal{L}_{se}$}  & \multicolumn{4}{c}{\textbf{Performance}} \\ 
\cmidrule{5-8}
                             &            &            &             & R@0.3 & R@0.5 & R@0.7 & mIoU  \\ \midrule
\multirow{5}{*}{\rot{\parbox[c]{2cm}{\centering Transformer-base}}} &       &            &             & \OOM  & \OOM  & \OOM  & \OOM   \\
                             & \cellcolor[HTML]{EFEFEF}\checkmark &  \cellcolor[HTML]{EFEFEF}          &  \cellcolor[HTML]{EFEFEF}           & \cellcolor[HTML]{EFEFEF}33.65 & \cellcolor[HTML]{EFEFEF}20.50 & \cellcolor[HTML]{EFEFEF}9.05  &\cellcolor[HTML]{EFEFEF} 22.23 \\
                             & \checkmark &            & \checkmark  & 33.71 & 19.90 & 9.08  & 21.89 \\
                             & \cellcolor[HTML]{EFEFEF}\checkmark & \cellcolor[HTML]{EFEFEF}\checkmark &  \cellcolor[HTML]{EFEFEF}           & \cellcolor[HTML]{EFEFEF}40.86 & \cellcolor[HTML]{EFEFEF}25.09 & \cellcolor[HTML]{EFEFEF}11.51 & \cellcolor[HTML]{EFEFEF}26.16 \\
                             & \checkmark & \checkmark & \checkmark  & 41.75 & 26.52 & 13.34 & 27.83 \\ \midrule
\multirow{5}{*}{\rot{\parbox[c]{1cm}{\centering BERT-base}}}   &   \cellcolor[HTML]{EFEFEF}         &  \cellcolor[HTML]{EFEFEF}          &   \cellcolor[HTML]{EFEFEF}          & \cellcolor[HTML]{EFEFEF}\OOM  & \cellcolor[HTML]{EFEFEF}\OOM  & \cellcolor[HTML]{EFEFEF}\OOM  & \cellcolor[HTML]{EFEFEF}\OOM   \\
                             & \checkmark &            &             & 37.29 & 22.51 & 10.74 & 24.61 \\
                             & \cellcolor[HTML]{EFEFEF}\checkmark & \cellcolor[HTML]{EFEFEF}           & \cellcolor[HTML]{EFEFEF}\checkmark  & \cellcolor[HTML]{EFEFEF}39.46 & \cellcolor[HTML]{EFEFEF}25.60 & \cellcolor[HTML]{EFEFEF}12.57 & \cellcolor[HTML]{EFEFEF}26.41 \\
                             & \checkmark & \checkmark &             & 41.32 & 27.38 & 14.15 & 27.88 \\
                             & \cellcolor[HTML]{EFEFEF}\checkmark & \cellcolor[HTML]{EFEFEF}\checkmark & \cellcolor[HTML]{EFEFEF}\checkmark  & \cellcolor[HTML]{EFEFEF}42.18 & \cellcolor[HTML]{EFEFEF}28.01 & \cellcolor[HTML]{EFEFEF}14.63 & \cellcolor[HTML]{EFEFEF}28.24 \\\midrule
\multirow{5}{*}{\rot{\model{}}}    &            &            &             & \OOM  & \OOM & \OOM  & \OOM   \\
                             & \cellcolor[HTML]{EFEFEF}\checkmark &  \cellcolor[HTML]{EFEFEF}          &   \cellcolor[HTML]{EFEFEF}          & \cellcolor[HTML]{EFEFEF}42.33 & \cellcolor[HTML]{EFEFEF}28.26 & \cellcolor[HTML]{EFEFEF}12.83 & \cellcolor[HTML]{EFEFEF}27.49 \\
                             & \checkmark &            & \checkmark  & 42.84 & 28.01 & 13.80 & 28.04 \\
                             & \cellcolor[HTML]{EFEFEF}\checkmark & \cellcolor[HTML]{EFEFEF}\checkmark & \cellcolor[HTML]{EFEFEF}            & \cellcolor[HTML]{EFEFEF}46.39 & \cellcolor[HTML]{EFEFEF}30.58 & \cellcolor[HTML]{EFEFEF}15.35 & \cellcolor[HTML]{EFEFEF}30.57 \\
                             & \checkmark & \checkmark & \checkmark  & \textbf{46.76} & \textbf{31.33} & \textbf{15.81} & \textbf{30.92} \\ \bottomrule % 9c7c
\end{tabular}
    \caption{Results of our Transformer ablation study, performed on the YouCookII dataset, where \OOM{} indicates we obtained an out-of-memory error even when using the largest GPU at our disposal. }
    \vspace{-0.5cm}
    \label{table:ablations}
\end{table}

\begin{table*}[ht!]
    \centering
    \rowcolors{5}{gray!15}{white}
    \scalebox{0.9}{
    \begin{tabular}{lcccccccccccc }
        \toprule
        \multirow{2}{*}{\textbf{Method}} & \multicolumn{4}{c}{\textbf{Charades-STA}} & \multicolumn{4}{c}{\textbf{ActivityNet}} & \multicolumn{4}{c}{\textbf{YouCookII}}\\
        \cmidrule{2-13}
         & R@0.3 & R@0.5 & R@0.7 & mIoU & R@0.3 & R@0.5 & R@0.7 & mIoU & R@0.3 & R@0.5 & R@0.7 & mIoU \\
        \midrule
        Random & - & 8.51 & 3.03 & - & 5.60 & 2.50 & 0.80 & - & 4.84 & 1.72 & 0.60 & -  \\
        CTRL & - & 21.42 & 7.15 & - & 28.70 & 14.00 & - & 20.54 & - & - & - & -  \\
        ABLR $\dag$ & - & 24.36 & 9.00 & - & 55.67 & 36.79 & - & 36.99 & - & - & - & - \\
        TripNet & 51.33 & 36.61 & 14.50 & - & 48.42 & 32.19 & 13.93 & - & - & - & - & -  \\
        CBP & 50.19 & 36.80 & 18.87 & 35.74 & 54.30 & 35.76 & 17.80 & 36.85 & - & - & - & -  \\
        MAN & - & 46.53 & 22.72 & - & - & - & - & - & - & - & - & - \\
        ExCL $\ddag$ & 65.10 & 44.10 & 22.60 & - & 62.10 & 41.60 & 23.90 & - & 26.58 & 15.72 & 8.19 & 18.99 \\
        TMLGA & 67.53 & 52.02 & 33.74 & 48.22 & 51.28 & 33.04 & 19.26 & 37.78 & 33.48 & 20.65 & 10.94 & 23.07 \\
        LGVTI & 72.96 & 59.46 & 35.48 & 51.38 & 58.52 & 41.51 & 23.07 & 41.13 & - & - & - & - \\ % 
        DORi & 72.72 & 59.65 & \bf40.56 & \bf53.28 & 57.89 & 41.49 & 26.41 & 42.78 & 43.36 & 30.47 & \bf18.24 & 30.46 \\
        VSLNet & 70.46 & 54.19 & 35.22 & 50.02 & \bf63.16 & 43.22 & 26.16 & 43.19  & - & - & - & - \\
        CPNet & - & \bf60.27 & 38.74 & 52.00 & - & 40.56 & 21.63 & 40.65 & - & - & - & -  \\
        CPN & \bf75.53 & 59.77 & 36.67 & 53.14 & 62.81 & 45.10 & 28.10 & \bf45.70 & - & - & - & - \\
        % Multi-stage Aggregated Transformer
        MSAT & - & - & - & - & - & \bf48.02 & \bf31.78 & - & - & - & - & - \\
       %  LaCoSTe {\tiny ECCV} & - & - & - & - & - & - & - & - & 42.70 & 29.04 & 16.29  \\ 
        \midrule
        % \model{} & 71.88 & 58.52 & 38.51 & 51.76 & \textbf{59.83} & \textbf{42.38} & \textbf{25.93} & \textbf{43.19}
        \model{} & 71.88 & 58.52 & 38.51 & 51.76 & 60.61 & 43.74 & 27.04 & 44.05 & \bf46.76 & \bf31.33 & 15.81 & \bf30.92 \\
        % \model{} & & & & & & & & \\
        \bottomrule
    \end{tabular}
    }
    \caption{Performance comparison of our approach with existing methods for different tIoU $\alpha$ levels. Values are reported on the validation split of Charades-STA and the ActivityNet Captions. $\dag$ Results for ABLR are as reported by \cite{sap2019}. $\ddag$ The results reported by ExCL for ActivityNet have 3,370 missing videos, and the results on YoucookII were obtained using our own implementation.}
    \label{table:sota}
    \vspace{-0.5cm}
\end{table*}

\begin{table}[t]
    \centering
    % \rowcolors{5}{gray!15}{white}
    \footnotesize
\begin{tabular}{cl@{\hspace{0.12cm}}c@{\hspace{0.1cm}}cccc}
\toprule
 &  & Memory & \multicolumn{4}{c}{Performance} \\
\multirow{-2}{*}{\bf Data} & \multirow{-2}{*}{\bf Model} & GB & R@0.3 & R@0.5 & R@0.7 & mIoU \\ 
 \midrule
 \multirow{6}{*}{\rot{\textbf{Charades-STA}}} & ExCL & 2.8 & 62.28 & 39.73 & 22.53 & 42.28 \\
 & \cellcolor[HTML]{EFEFEF}+ \sampling{} & \cellcolor[HTML]{EFEFEF}1.6 & \cellcolor[HTML]{EFEFEF}\textbf{62.74} & \cellcolor[HTML]{EFEFEF}\textbf{42.04} & \cellcolor[HTML]{EFEFEF}\textbf{24.57} & \cellcolor[HTML]{EFEFEF}\textbf{43.05} \\
 \cmidrule{2-7}
 & TMLGA & 2.3 & 67.53 & 52.02 & 33.74 & 48.22 \\
 & \cellcolor[HTML]{EFEFEF}+ \sampling{} & \cellcolor[HTML]{EFEFEF}1.5 & \cellcolor[HTML]{EFEFEF}\textbf{70.67} & \cellcolor[HTML]{EFEFEF}\textbf{52.20} & \cellcolor[HTML]{EFEFEF}\textbf{33.90} & \cellcolor[HTML]{EFEFEF}\textbf{49.18} \\
 \cmidrule{2-7}
 & DORi$^{\clubsuit}$ & 32.8& 72.72 & 59.65 & 40.56 & 53.28 \\
 & \cellcolor[HTML]{EFEFEF}+ \sampling{} & \cellcolor[HTML]{EFEFEF}23.1 & \cellcolor[HTML]{EFEFEF}\textbf{72.90} & \cellcolor[HTML]{EFEFEF}\textbf{59.67} & \cellcolor[HTML]{EFEFEF}\textbf{40.94} & \cellcolor[HTML]{EFEFEF}\textbf{53.44} \\ 
 \midrule
 \multirow{6}{*}{\rot{\textbf{ActivityNet}}}& ExCL & 6.4 & 55.49 & 39.33 & 24.04 & 40.32 \\
 & \cellcolor[HTML]{EFEFEF}+ \sampling{} & \cellcolor[HTML]{EFEFEF}1.8 & \cellcolor[HTML]{EFEFEF}\textbf{56.42} & \cellcolor[HTML]{EFEFEF}\textbf{40.37} & \cellcolor[HTML]{EFEFEF}\textbf{24.70} & \cellcolor[HTML]{EFEFEF}\textbf{41.13} \\
 \cmidrule{2-7}
 & TMLGA & 7.3 & 51.28 & 33.04 & 19.56 & 37.78 \\
 & \cellcolor[HTML]{EFEFEF}+ \sampling{} & \cellcolor[HTML]{EFEFEF}1.7 & \cellcolor[HTML]{EFEFEF}\textbf{53.00} & \cellcolor[HTML]{EFEFEF}\textbf{35.10} & \cellcolor[HTML]{EFEFEF}\textbf{19.83} & \cellcolor[HTML]{EFEFEF}\textbf{37.85} \\
 \cmidrule{2-7}
 & DORi$^{\clubsuit}$ & 34.7 & 57.89 & 41.35 & \textbf{26.41} & 42.79 \\
 & \cellcolor[HTML]{EFEFEF}+\sampling & \cellcolor[HTML]{EFEFEF}24.0 & \cellcolor[HTML]{EFEFEF}\textbf{58.89} & \cellcolor[HTML]{EFEFEF}\textbf{42.21} & \cellcolor[HTML]{EFEFEF}26.36 & \cellcolor[HTML]{EFEFEF}\textbf{43.02} \\ 
 \midrule
 \multirow{6}{*}{\rot{\textbf{YouCookII}}} & ExCL & 6.9 & 26.58 & 15.72 & 8.19 & 18.99 \\
 & \cellcolor[HTML]{EFEFEF}+ \sampling{} & \cellcolor[HTML]{EFEFEF}1.8 & \cellcolor[HTML]{EFEFEF}\textbf{30.96} & \cellcolor[HTML]{EFEFEF}\textbf{18.64} & \cellcolor[HTML]{EFEFEF}\textbf{10.05} & \cellcolor[HTML]{EFEFEF}\textbf{21.76} \\
 \cmidrule{2-7}
 & TMLGA & 9.5 & 33.48 & 20.65 & 10.94 & 23.07 \\
 & \cellcolor[HTML]{EFEFEF}+ \sampling{} & \cellcolor[HTML]{EFEFEF}1.8 & \cellcolor[HTML]{EFEFEF}\textbf{39.29} & \cellcolor[HTML]{EFEFEF}\textbf{25.40} & \cellcolor[HTML]{EFEFEF}\textbf{12.80} & \cellcolor[HTML]{EFEFEF}\textbf{26.20} \\
 \cmidrule{2-7}
 & DORi$^{\clubsuit}$ & 46.4 & 43.36 & 30.47 & 18.24 & 30.46 \\
& \cellcolor[HTML]{EFEFEF}+ \sampling & \cellcolor[HTML]{EFEFEF} 24.2 & \cellcolor[HTML]{EFEFEF}\textbf{46.74} & \cellcolor[HTML]{EFEFEF}\textbf{32.19} & \cellcolor[HTML]{EFEFEF}\textbf{18.33} & \cellcolor[HTML]{EFEFEF}\textbf{31.69} \\ \bottomrule
\end{tabular}
    \caption{Results of our experiments combining \sampling{} with existing work. Except where indicated, experiments were performed using a batch size 32. $^\clubsuit$ indicates experiments performed using a batch size of 4 due to memory constraints.}
    % which we needed to use for DORi which otherwise would does not fit in memory
    \label{table:sampling_generalization}
    \vspace{-2em}
\end{table}

%  the TMLGA and DORi models. \YM{why not show the result of your transformer model in this table?  it should be shown (and compared with other methods) before ablation tests only this table lists methods in a horizontal way, while others vertically.  it's better if this table also can do it vertically.}

Next, we study the impact of \sampling{} at different bucket sizes. For these experiments we use the YouCookII dataset, which contains the longest videos on average, and test bucket sizes ranging from 100 to 500. As we can see on Table \ref{table:bucket_size_study}, variations on parameter $b$ have an impact consistent with its expected behavior, with diminishing results as $b$ increases, and a clear performance sweet-spot at $b=200$ which we adopt for the rest of the experiments in this paper.

Finally, we ablate \model{} component-by-component, comparing it with two highly-competitive Transformer-based model variations: (1) \textbf{Transformer-base} a randomly-initialized multi-modal Transformer-base\footnote{We follow the original notation \cite{Vaswani_NIPS_2017} using 12 layers and 12 attention heads.}, into which we directly feed the text input and the sampled video features, previously embedding them using a learned embedding matrix and a linear projection layer, respectively. Each encoder uses a separate set of positional embeddings, and we also add a type embedding to indicate the model the nature of each vector. After this, the embedded sequences are concatenated lengthwise and passed through the transformer blocks; (2) \textbf{BERT-base}, where we initialize our Transformer-base variation with the weights of BERT. In this case, the projection linear layer of the video features and the respective positional encodings are randomly initialized.

Both model variations apply our attention guiding loss $\mathcal{L}_{att}$ to all the attention heads in all the layers. This effectively means that there is no functionality separation inside these models. We note that these baselines are comparable to existing multi-modal transformer models such as VilBERT \cite{luViLBERTPretrainingTaskAgnostic2019} and UNITER \cite{chenUNITERUNiversalImageTExt2020}. Experiments are again performed in YouCookII, which contains the longest videos.

As Table \ref{table:ablations} shows, we see that \model{} is able to consistently outperform our Transformer-based variations, with each ablated component clearly contributing to increase performance. The results also reflect the importance of \sampling{}, enabling all kinds of Transformer-based models we tested to process long untrimmed videos, which would otherwise lead to out-of-memory errors. 

Regarding the interaction of the attention loss with different model variations, we see that this additional training signal leads to consistent gains for all Transformer-based variations, but that these are larger in the case of \model{}. We surmise this is due to the attention loss potentially interfering with the inductive bias that the baselines require to process the multi-modal inputs, as well as with the already acquired bias in the case of BERT, reflected in certain attention patterns for each head which have been studied and documented by Rogers \etal. \cite{rogers-etal-2020-primer} among others. This ultimately highlights the importance of separating functionality inside Transformer models for our task, which allows our model to perform better overall.

\subsection{Comparison to state-of-the-art models}

We compare the performance of \model{} on the datasets considered against several prior work, as well as to a random baseline that simply selects an arbitrary video segment as the moment for each example. We consider a broad selection of models based on different approaches, specifically proposal-based techniques including CTRL \cite{Gao_2017_ICCV}, MAN \cite{zhang2018man}, CBP \cite{wang2019temporally} and the more recent multi-stage Transformer approach (MSAT) by Zhang \etal. \cite{zhangMultiStageAggregatedTransformer2021}, as well as TripNet \cite{hahn2019tripping}, a method based on reinforcement learning. In addition to that, we also compare our approach to more recent methods that do not rely on proposals, including ABLR \cite{yuan2018find}, ExCL \cite{ghosh2019excl}, TMLGA \cite{rodriguez2019proposal} and LGVTI \cite{munLocalGlobalVideoTextInteractions2020}, as well as more recent approaches including CPNet \cite{liProposalFreeVideoGrounding2021}, VSLNet \cite{zhangSpanbasedLocalizingNetwork2020a}, CPN \cite{zhaoCascadedPredictionNetwork2021} and DORi \cite{rodriguez-opazoDORiDiscoveringObject2021}. These two last models contain specifically-crafted graph-based approaches for the task, with DORi also incorporating spatio-temporal features.

% Additionally, we also consider DORi \cite{rodriguez-opazoDORiDiscoveringObject2021} another proposal-free model which incorporates spatio-temporal features and whose implementation we also directly integrate into our codebase. 

% Although DORi is not directly comparable to other models since it relies on additional spatial information extracted from the video, its performance can could be regarded as an upper bound for the rest of the experiments on this paper.

Table \ref{table:sota} summarizes our best results on Charades-STA, ActivityNet Captions and YouCookII datasets, while also comparing the obtained performance to relevant prior work. We can see that overall \model{} is able to offer excellent performance, closing the gap with sophisticated graph-based models like CPN and DORi, and obtaining a new sate-of-the art in YouCookII, showing the effectiveness of our approach when dealing with long untrimmed videos.

%It is possible to see that our method is able to outperform previous work by a consistent margin, specially for the $\alpha=0.7$ band and also in terms of the mean tIoU (mIoU).

\subsection{Combining \sampling{} with previous work}
% \EMT{Find a better name for this subsection}
\label{sec:sampling_combined}
% and its sensitivity to changes in the bucket size.
We finally focus on studying the ability of our sampling technique to be combined with different models. We do this by incorporating \sampling{} into three proposal-free models selected from the literature, and testing them on our datasets. We consider ExCL \cite{ghosh2019excl} and TMLGA \cite{rodriguez2019proposal}, which have been extensively studied in the past years, as well as DORi. We utilized our own implementation of ExCL with our I3D features extracted at 25 fps\footnote{The original implementation sampled at a frame-rate of 5 fps.}, and directly integrated the original implementations of the latter into our code.

As Table \ref{table:sampling_generalization} shows, \sampling{} is able to consistently provide performance improvements in all cases, with gains of up to 3.13\% in terms of the mean temporal IoU. These improvements lead to new state-of-the-art results on both the Charades-STA and YouCookII datasets. We note that despite not having access to the spatial information that DORi incorporates, the performance of \model{} is very competitive to that of DORi+\sampling{} in YouCookII, which contains the longest videos. This again illustrates the effectiveness of our approach in this scenario.

% \subsection{Qualitative Analysis}

\subsection{Limitations and Societal Impact}
Although our sampling technique has shown to be very efficient and effective in different datasets, there is a specific scenario where it could degrade the performance of a given model. This scenario occurs when the span of the query, \textit{a.k.a.} moment, is located completely inside a given bucket, with many additional frames/features in the same bucket. In this case, the best that a model can do is that predict the moment happens inside of that bucket, losing finer granularity. In practice, this occurs then when the ratio between the duration of a given moment and the duration of the video is vanishingly small.
% We believe this issue can be alleviated in future work by recursively generating and exploring buckets.
In terms of societal impact, models that excel at temporal moment localization could be used to break private information and manipulate people's behavior. On the other side, the benefit of less memory usage reduces GPU power consumption and therefore could facilitate access to tasks related to videos for researchers with fewer resources.

\section{Conclusion}

In this paper we have presented \model{}, a Transformer-based model for the task of temporal moment localization which operates at a constant maximum memory footprint regardless of the input length. The success of our model fundamentally relies on our modular design, which allows us to separate functionality, and \sampling{}, where we split the sequence of input video features into a fixed number of buckets and select a single feature per bucket using a stochastic approach. Experiments conducted on three challenging datasets show that \model{} obtains excellent results, being able to obtain state-of-the-art performance on YouCookII. We also show that our sampling technique can improve the performance of prior work on all considered datasets, leading to a new state-of-the-art on Charades-STA. We think these results highlight the importance of sampling techniques as a valid mechanism to obtain better coverage of long input videos while keeping memory usage low.
% \mr{i honestly have no idea what this ``under budget'' means :') maybe better to put it in more specific terms---(bc people's budgets are different etc...)}. 

% This ultimately provides a concrete direction for further research on tasks where it is necessary to cover long untrimmed videos, which include but are not limited to video grounding. 

For future work, we are interested in testing our sampling-based approach in other relevant tasks in the context of video-and-language, for example as video retrieval. We are also interested in extending our approach to address its limitations, for example, using adaptive or iterative sampling to treat different areas of the video with different granularity as required.

% \BF{Have we considered sampling K frames from a bucket and then applying max/average pooling}
% \EMT{I discussed a similar idea with Chis a while a go, but I haven't tried it yet.}

% \clearpage
% \newpage
{\small
\bibliographystyle{ieee_fullname}
\bibliography{bibliography}
}
% \clearpage
\newpage
\appendix
% \section*{Appendices}
\section{Sufficient Statistic}
A \textit{statistic} is a function $T = r(X_1, \ldots, X_n)$ of the random sample $X_1, \ldots, X_n$, which carries information of the sampled data, such as the sample mean and sample variance. We say that a statistic satisfies the criterion of sufficiency when no other statistic which can be calculated from the same sample provides any additional information as to the value, of the parameter to be estimated. We can easily find a sufficient statistics by using the \textit{Fisher–Neyman Factorization Theorem}.

\noindent \textit{\textbf{Factorization theorem}}: 
\textit{given a random sample $X_1, \ldots, X_n$ with joint density $f(x_1, \ldots, x_n | \theta)$ a \textit{statistic} $T = r(X_1, \dots, X_n)$ is sufficient if and only if the joint density can be factored as follows:
\begin{align*}
    f(x_1,\ldots, x_n | \theta) =& u(x_1, \ldots, x_n) v(r(x_1, \ldots, x_n),\theta) 
\end{align*}
where $u$ and $v$ are non-negative functions. The function $u$ can depend on the full random sample $x_1,\dots,x_n$ but not on the unknown parameter $\theta$. The function $v$ can depend on $\theta$, but can depend on the random sample only through the value of $r(x_1,\dots, x_n)$.}

In our case, let us assume that our bucket contains features $X_i$ that are independent and uniformly distributed on $[0, \theta]$ where $\theta$ is unknown. Then, the probability dense function can be written as a product of individual densities since the observations are independent,
\begin{align*}
    f (x_1 , \ldots , x_n|\theta) =& \frac{1}{\theta}\textbf{1}_{\{0 \leq x_1 \leq \theta\}} \dots \frac{1}{\theta}\textbf{1}_{\{0 \leq x_n \leq \theta\}}
    % f(x_1, \dots, x_n |\theta) =& \frac{1}{\theta^{n}} \textbf{1}(0 \leq x_i \leq \theta, i = 1, \dots, n)
\end{align*}
Here $\textbf{1}(E)$ is an indicator function. It is $1$ if the event $E$ holds, and $0$ if it does not. Now $x_i \leq \theta$ for $i = 1, \dots , n$ if and only if $\text{max}\{x_1, \dots, x_n\} \leq \theta$. Therefore,
\begin{equation*}
    % f (x_1 , \dots , x_n|\theta) = \frac{1}{\theta^{n}} \textbf{1}(\text{max}\{x_1 , \dots , x_n \} \leq \theta)\textbf{1}(0 \leq \text{min}\{x_1 , \dots , x_n \})
    f (x_1, \ldots, x_n|\theta) = \frac{1}{\theta^n}\textbf{1}_{\{0 \leq \min{\{x_i\}\}}}\textbf{1}_{\{\max{\{ x_i\} \leq \theta\}}}
\end{equation*}
Thus, the \textit{factorization theorem} shows that $T = \max\{X_1, \ldots, X_n \}$ is a sufficient statistic since the density function takes the required form, where $u=\textbf{1}_{\{0 \leq \min\{x_i\}\}}$ and $v = \frac{1}{\theta^n}\textbf{1}_{\{\max{\{ x_i\} \leq \theta\}}}$, which is a function that only depends on $\theta$ and $T=\max\{x_i\}$.

\section{Dynamic-Rate Feature Down-Sampling}

In this section, we present details of our Dynamic-Rate Feature Down-Sampling ablation experiments. With this sampling heuristic, our intention was to create buckets that satisfy the two following conditions. First, we would like each bucket to hold semantically similar features, using similarity on the embedding space as a proxy. Second, we aim to group features in a way such that the number of buckets $l$ that hold the totality of the features in the video is smaller than the desired number of buckets $b$. 

The bucket construction procedure works as follows. For an input feature sequence of length $n$, we use the cosine distance to compute the semantic similarity between each of the features in the video, and construct the pairwise distance matrix $\mD \in \sR^{n \times n}$. We then start the process with a single bucket that contains only the first feature $x_1$, and add features to this bucket starting from $x_2$. Feature $x_2$ will be added to the bucket if and only if the cosine-distance $D_{1, 2} < th$, where $th$ is a threshold parameter, and otherwise a new bucket is started and the process is repeated until all the features have been processed. Once this is done, we evaluate the number of buckets $l$ that were created, and if $l > b$, then we reduce the threshold $th$ by a small margin ($0.01$) and generate all the buckets again. This process is repeated until the $l \leq b$ condition is satisfied. Please see Algorithm \ref{alg:cap} below, for additional details.

\section{Qualitative Results}

\begin{figure*}[t]
    \centering
    \includegraphics[width=0.9\linewidth]{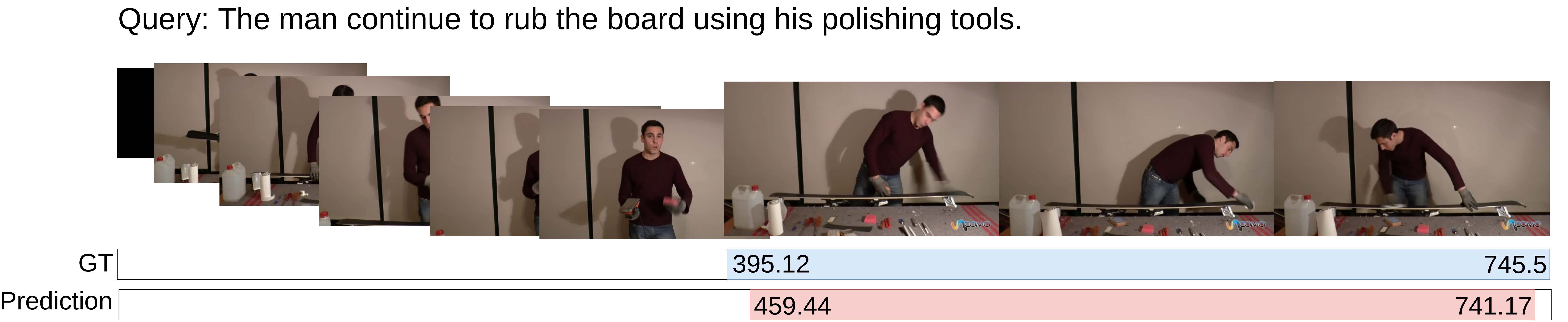}
    \caption{Results of our method on the ActivityNet dataset in a very long video of 12 minutes and 25 seconds (745.5 seconds). Our method can localize the query \textit{The man continue to rub the board using his polishing tools} with a temporal IoU of 80.41\%}
    \label{fig:anetq}
\end{figure*}

\begin{figure*}[t]
    \centering
    \includegraphics[width=0.9\linewidth]{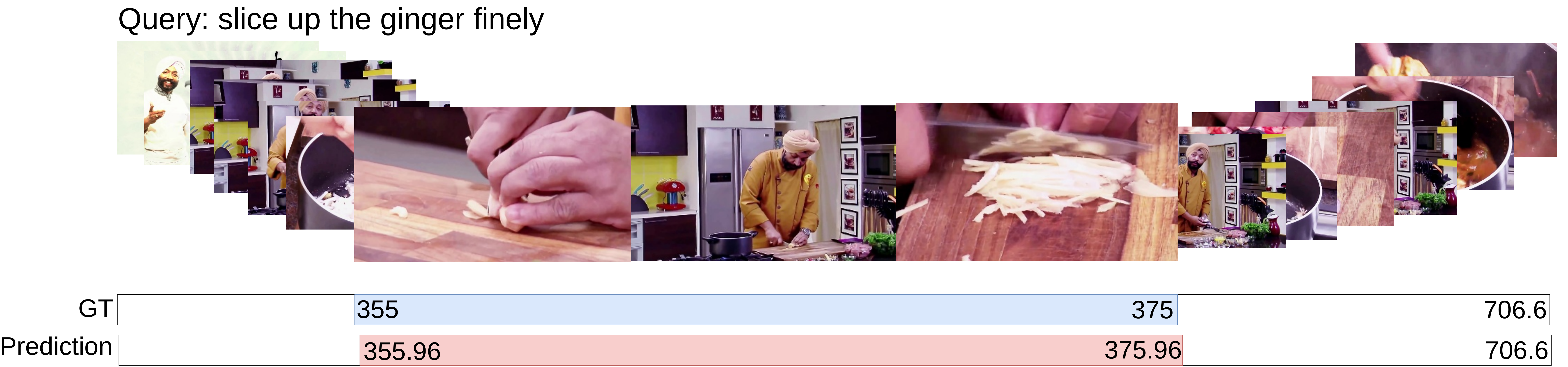}
    \caption{Qualitative result of our method on the YouCooKII dataset in one of the longest videos in the dataset (11 minutes and 46 seconds.) Our method obtains a temporal IoU of 90.83\% for the query \textit{slice up the ginger finely}.}
    \label{fig:youcookIIq}
\end{figure*}

\begin{figure*}[t]
    \centering
    \includegraphics[width=0.9\linewidth]{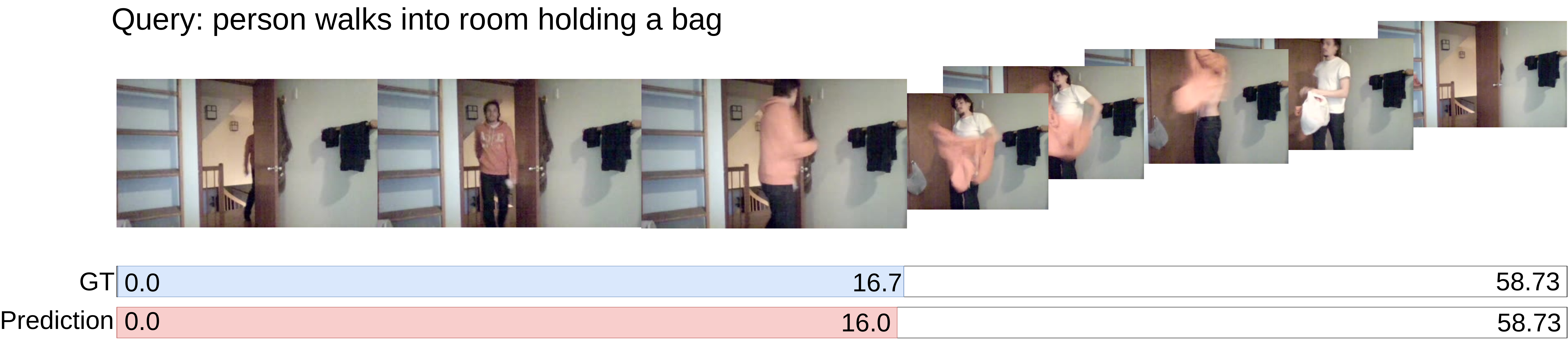}
    \caption{Charades-STA qualitative results on one of the longest videos in the dataset. For the query  \textit{person walks into room holding a bag}, our method obtains 95.70\% IoU with respect to the ground truth annotations.}
    \label{fig:charadesq}
\end{figure*}

In this section, we present qualitative results of our method for each one of datasets we use for evaluation. Ground truth (GT) and predictions in Figures \ref{fig:anetq}, \ref{fig:youcookIIq} and \ref{fig:charadesq} are in seconds.

\begin{algorithm}[H]
\caption{Dynamic-rate Feature down-sampling using cosine similarity Algorithm}
\label{alg:cap}
\begin{algorithmic}

% \Require $n \geq 0$
% \Ensure $y = x^n$
\State $\mD = 1 - \text{pairwise\_distances}(V)$
\State $th = 1.0$
\State $flag = True$
% \State $X \gets x$
% \State $N \gets n$
\State index\_sample = \{\}
\While{$flag$}
    \State indx $= 0$
    \State $st = 0$
    \State $ed = 1$
    \For{$ed \gets st$ to len($V$)} 
        \State $s = \mD_{st,ed}$
        \If{$s < th$}
            \State samples\_in\_bucket = []
            \For{$i \gets st$ to $ed$}
                \State samples\_in\_bucket.append(i)
            \EndFor
            \State index\_sample[indx] = samples\_in\_bucket
            \State $st = ed$
            \State indx = indx + 1
        \EndIf
        \If{indx $<=$ bucket\_size}:
            \State flag = False
        \Else
            \State $th = th - 0.01$
            \State index\_sample = \{\}
        \EndIf
    \EndFor
\EndWhile
\end{algorithmic}
\end{algorithm}
As seen on Figure \ref{fig:anetq}, in the case of ActivityNet Caption, our method is able to localize the query \textit{The man continue to rub the board using his polishing tools} with a high temporal intersection over union (IoU) of 80.41\%. Though not visible in the figure, we also note that the end of this video is full of black frames and information about the creator, \eg, webpage and logos. This exemplifies how ground truth annotations can be inaccurate, and how our model can adequately deal with these issues.

Figure \ref{fig:youcookIIq} presents qualitative results for YookCookII dataset. In this case, we specifically  present our predictions on one of the longest videos in the dataset, with a duration of 11 minutes and 46 seconds, and where the natural language query is \textit{slice up the ginger finely}. As seen, our method obtains an impressive performance considering that the moment of interest lasts only 20 seconds. This figure also serves to exemplify one of the limitations of the bucketing approach we take. It is possible to see that the predictions of our model, though precise overall, add 0.96 seconds to both the start and end locations. This is a result of the maximum granularity given by the buckets and features in our system.

Finally, Figure \ref{fig:charadesq} shows an example of the predictions of our model on the Charades-STA dataset. In this case, we also choose one of the longest videos in the data, with a duration of approximately 1 minute. For the query \textit{person walks into room holding a bag}, our method obtains a good performance of 95.70\% of temporal IoU.

% This is repeated until the cwhere
%th = 1.0
% while flag:
%     indx = 0
%     st = 0
%     ed = 1
%     index_sample = {}
%     for ed in range(feature.shape[0]):
%         cos_sim = dist_out[st, ed]
%         if cos_sim < th:
%             samples_in_bucket = []
%             for i in range(st, ed):
%                 samples_in_bucket.append(i)
%             print(indx, st, ed, ed - st, cos_sim, th)
%             index_sample[indx] = samples_in_bucket
%             st = ed
%             indx += 1
%     if indx <= args.bucket_size:
%         flag = False
%     else:
%         th -= 0.01

\end{document}